\documentclass[runningheads]{llncs}
\usepackage[T1]{fontenc}
\usepackage{graphicx}
\usepackage{xcolor}
\usepackage{tcolorbox}
\usepackage{adjustbox}
\usepackage{listings}
\usepackage{subcaption}
\usepackage{ifthen}
\usepackage{graphicx}
\usepackage{subcaption}

\lstdefinestyle{grammar}{
    basicstyle=\ttfamily\small
}

\usepackage{stix}

\newboolean{showcomments}
\setboolean{showcomments}{true}
\ifdefined\notodocomments
  \setboolean{showcomments}{false}
\fi
\ifthenelse{\boolean{showcomments}}
 { \newcommand{\mynote}[2]{
      \fbox{\bfseries\sffamily\scriptsize#1}
        {\small$\blacktriangleright$\textsf{\emph{#2}}$\blacktriangleleft$}}}
        { \newcommand{\mynote}[2]{}}

\begin{document}
\title{Contribution Title}
\title{SecretFan: Synthesizing Realistic Data without~Breaking~Privacy}
\author{Laura Plein\inst{1} \and Alexi Turcotte\inst{1} \and Arina Hallemans\inst{2} \and Andreas Zeller\inst{1}}
\institute{CISPA Helmholtz Center for Information Security \and Saarland University}
%\author{Alexi Turcotte}
%\institute{CISPA Helmholtz Center for Information Security}
%\author{Arina Hallemans}
%\institute{Saarland University}
%\author{Andreas Zeller}
%\institute{CISPA Helmholtz Center for Information Security}
%\author{First Author\inst{1}\orcidID{0000-1111-2222-3333} \and
%Second Author\inst{2,3}\orcidID{1111-2222-3333-4444} \and
%Third Author\inst{3}\orcidID{2222--3333-4444-5555}}
%\institute{Princeton University, Princeton NJ 08544, USA \and
%Springer Heidelberg, Tiergartenstr. 17, 69121 Heidelberg, Germany
\maketitle              % typeset the header of the contribution
\begin{abstract}
\iffalse
Machine learning models are increasingly applied in various sensitive domains such as health,
finance, and insurance. They often need to be trained on data that contains personally identifiable
information or sensitive attributes, which raises privacy concerns. In critical domains, it is also essential
to thoroughly test models under realistic conditions to ensure their reliability and robustness. However,
testers typically do not have access to the original datasets, since these are usually restricted due to
confidentiality and data protection requirements.
\fi
There is a need for synthetic training and test datasets that replicate statistical distributions of original datasets without compromising their confidentiality.
A lot of research has been done in leveraging  Generative Adversarial Networks (GANs) for synthetic data generation, however the resulting models are either not accurate enough or are still vulnerable to membership inference attacks (MIA) or dataset reconstruction attacks since the original data has been leveraged in the training process.

In this paper, we frame synthetic data generation as a guided test generation, or search-based testing problem rather than a purely generative modeling task.
Ours is a search-based, adequacy-guided input generation technique inspired by GANs, with a \textit{generation} step and a \textit{discrimination} step; as in GAN, \textit{discrimination} uses a discriminator model trained on the date, but instead of using models also for generation, we use a fuzzer.
This way, the original (private) data is only indirectly leveraged in the generation process, and by evolving samples and determining ``good samples'' with the discriminator, we can generate privacy-preserving data that follows the same statistical distributions as the original dataset, leading to a similar utility as the original data. 
We evaluated our approach on eight datasets that have been used to evaluate the state-of-the-art techniques, finding that synthetic generated with our technique achieves good utility on average while also having good similarity scores, highlighting the potential of a mixed approach leveraging classical generation and model-driven discrimination for generating privacy-preserving, useful synthetic datasets.

\keywords{Synthetic Data Generation  \and Model Training \& Validation \and Privacy Preserving}
\end{abstract}

\section{Introduction}

In the past years, Artificial Intelligence (AI) systems have become increasingly prominent in sensitive domains such as health~\cite{liu2024preserving,hernandez2022synthetic}, finance~\cite{assefa2020generating}, and education~\cite{liu2024scaling}.
Leveraging AI systems in such domains requires thorough testing and validation---especially since they are typically trained on highly personal information, which raises privacy concerns.
Specifically, AI trained in this way is vulnerable to Membership Inference Attacks (MIA) or dataset reconstruction attacks, which compromise the confidentiality of the original data. 

Generally, \emph{data sharing} is required for the training, testing, and validation of the AI system.
This, however, poses a problem: \emph{How can we prevent the leakage or memorization of private or confidential data?}
Clearly, you would not want some identifiable information about, say, your health records, to be suddenly reproduced by an AI system.
Over time, multiple techniques have been suggested to address this problem:

\begin{description}
\item[Anonymization.] To avoid the leakage of private data, the data is commonly anonymized before being shared.
While good in terms of privacy, this significantly compromises utility;
training a model where, e.g., the age of a person is replaced with an age range does provide better security, as it is harder to uniquely identify someone, but the model might miss important correlations between age and other features of the dataset.
Consequently, it's widely accepted that simply anonymizing the data is not an ideal solution for privacy-preserving AI systems~\cite{schneider2025data}.

\item[Synthetic data sets.] A lot of research has been done on \emph{synthesizing datasets} that can be used to train, test, and validate AI systems. 
In the past decade, Generative Adversarial Networks (GANs)~\cite{goodfellow2014generative} have gained a lot of attention due to their capabilities of generating synthetic data.
The main characteristic of GANs is that a discriminator and a generator compete against each other in order to generate quality data that cannot be distinguished from the original data. 
One of the main challenges with GANs is ``mode collapse''.
In most datasets, categorical features are typically heavily unbalanced. 
Thus, during the training of GANs, once the model detects that it can easily fool the discriminator by changing one of the features, it will typically focus on this particular feature and generate a huge amount of data with high similarity, failing to generalize beyond those samples~\cite{barsha2025depth}, leading to low utility. 

\item[Conditional generators.] In the context of generating privacy-preserving data, Xu et al.~\cite{xu2019modeling} introduce conditional GANs (CTGANs), which use a conditional generator to address these challenges.
In a CTGAN, conditions are extracted from the original distribution, and the model is trained by sampling, resulting in an even higher resemblance as the generator learns to make an exact copy of the original data.
A CTGAN is capable of generating synthetic data with satisfying utility, which can be great to increase the amount of available data, but the training process leverages the original data for training and focuses on high resemblance, so it remains vulnerable to data leakage, especially to membership inference attacks. CTGAN can be extended with differential privacy to reduce MIA attacks, however this comes at the cost of the data utility~\cite{hyeong2022empirical}. 

\end{description}

In summary, generating synthetic data that (a) provides strong privacy guarantees \emph{and}
(b)~captures the feature dependency of the original dataset remains a challenge, as also highlighted in this recent survey~\cite{challagundla2025synthetic}.
Hence, there is a clear need for more input diversity to achieve good utility while preserving privacy.

Now, generating bespoke data is not restricted to AI, and indeed the fields of search-based software testing, fuzzing, and adequacy-guided test generation have been thinking about how to generate input data for a long time; thus, in this paper, we frame synthetic data generation as a test generation problem, leveraging classical techniques for \textit{generation} and AI-based techniques for \textit{discrimination}.
Our approach is inspired by GAN networks, but we replace the generator with a \emph{fuzzer}---a random input generator, guided by a specification to produce useful and diverse inputs.
The goal is to generate the synthetic data without directly using the original data to provide better privacy guarantees.
Fuzzers are well known in the software engineering community to provide a huge amount of \emph{diverse} test inputs which are required to cover all the feature dependencies of the original dataset. However, a fuzzer by itself is not sufficient to produce data to train and test AI systems. Notably, common fuzzers lack the knowledge of how the different input features relate to each other.
Even specification-guided fuzzers, like \textit{grammar-based}~\cite{godefroid2008grammar} or \textit{language-based}~\cite{burkhardt1967generating,steinhofel2022input} fuzzers, which are at least syntax-aware generally have little-to-no semantic information, or the semantic information needs to be know a priori. 
Thus, samples generated with such a fuzzer would likely not follow the statistical distributions that are present between the specific features, nor the overall distribution of a specific feature throughout the dataset.
Therefore, we propose to combine the idea of GANs with a grammar-based fuzzer, by using a model to discriminate ``realistic'' generated inputs.
In our approach, we use the \textit{Fandango} fuzzer~\cite{zamudio2025Fandango} as a generator that allows us to leverage not only grammars but also introduce \textit{constraints} that generated inputs should satisfy.
Fandango \textit{satisfies} these constraints through \textit{evolutionary algorithms}, evolving a population of candidate inputs through grammar-aware mutations, and keeping those individuals that are closest to fulfilling the constraints.

The idea in this paper is to leverage feedback of a \textit{discriminator} to check whether the generated data is close to the original dataset or not; this provides insights into which generated samples are interesting.
Those \textit{interesting} samples are then returned to Fandango, which leverages them as an initial population for generating the next iteration of samples.
After performing several iterations of generating data with Fandango and adding the feedback of the discriminator in our loop, we can generate privacy-preserving data with satisfying utility.

The main contribution of our approach is to \emph{use a test generator to produce privacy-preserving data, where the generator has \emph{not} been trained on the original data, while providing sufficient diversity to unlikely leak any private or sensitive data from the original dataset.}
Additionally, our approach requires fewer resources than GAN-based solutions since our approach only requires a grammar as input and minimal computational resources, which allows us to quickly generate good samples.
To evaluate the feasibility and effectiveness of our approach, we assess the utility and resemblance of the generated data by training and testing models on the original and the generated data.
To confirm privacy, we verify first that our approach did not generate any samples that are present in the original dataset. Further, we perform a feature ablation study to identify in which scenarios partial samples can become exact matches of real records after suppressing a subset of features. Additionally, the statistical similarity of the synthetic and original datasets is assessed by computing the Jensen-Shannon divergence and Wasserstein distance. SecretFan\footnote{\url{https://github.com/LaPlei96/SecretFan}} and our experimental data are available on GitHub.

\section{Background}
\label{sec:background}

\subsection{Language-Based Testing}

In this paper, we leverage the Fandango language-based testing framework to generate rows of data.
In language-based testing~\cite{10.1145/3631520}, users specify a \textit{grammar} which describes the \emph{syntax} of inputs, as well as \textit{constraints} over the grammar which specify the \emph{semantics} of valid inputs.

\begin{figure}
    \centering
    \lstset{style=grammar}
    \begin{lstlisting}
<start>  ::= <header> '\n' <rows>
<header> ::= 'age' ', ' 'job' ', ' 'income' 
<rows>   ::= (<row> '\n')*
<row>    ::= <age> ', ' <job> ', ' <income> 
<age>    ::= <digit>+
<job>    ::= 'librarian' | 'neurosurgeon' | 'president' 
<income> ::= <digit>+
<digit>  ::= '0' | '1' | ... | '9'

where int(<age>) > 18 & int(<age>) < 70
    \end{lstlisting}
    \caption{Example CSV grammar and constraints.}
    \label{fig:background:example-grammar}
\end{figure}

\subsubsection*{Grammars.}
A context-free grammar (CFG) is defined as a 4-tuple $G = (V, \Sigma, R, S)$:
\begin{itemize}
    \item $V$ is a finite set of \textit{non-terminal} symbols, essentially variables in the grammar;
    \item $\Sigma$ is a finite set of \textit{terminal} characters, where $V \cap\Sigma=\emptyset$;
    \item $R$ is a finite relation $V \times(V \cup\Sigma)^*$ specifying the \textit{productions} or \textit{re-write rules} in the grammar, decribing how non-terminals can be \textit{expanded};
    \item $S$ is the start symbol, i.e., where the grammar begins.
\end{itemize}

For example, consider the grammar for a small comma-separated value file in Figure~\ref{fig:background:example-grammar}.
Here, \texttt{<start>} is the eponymous start symbol.
Symbols \texttt{*} is the Kleene star, which allows for zero or more repetitions: so, \texttt{(<row> '\textbackslash n')*} means zero or more \texttt{<row>} non-terminals separated by newlines.
We also use Kleene \texttt{+}, for one or more repetitions. 
The \texttt{|} symbol specifies options, so \texttt{<job>} can be either one of the alternatives.
A typical input that can be produced from this grammar would be:

\begin{verbatim}
age, job, income
29, librarian, 9427
78, president, 19300
\end{verbatim}

\subsubsection*{Constraints.}
In language-based testing, CFGs are augmented with \emph{constraints} over non-terminals.
We use Fandango~\cite{zamudio2025Fandango} in this paper, which has a small constraint language, but also allows users to specify additional constraints as Python functions. 
For example, take the \texttt{<age>} non-terminal from Figure~\ref{fig:background:example-grammar}---considering only the grammar, valid ages include \texttt{0}, \texttt{24}, and \texttt{75803812}, which is not realistic.
We can limit this with a \textit{constraint} over \texttt{<age>} by augmenting the grammar with a constraint as shown in the lower part of Figure~\ref{fig:background:example-grammar}; then, a line such as \texttt{78, president, 19300} would no longer be produced.
(We leave it to the reader to come up with realistic ranges for \texttt{income}.)

\subsubsection*{Fandango.}

As mentioned, we use the Fandango~\cite{zamudio2025Fandango} evolutionary language-based testing framework in this paper.
Being an \textit{evolutionary} approach, Fandango will begin by generating inputs according to the grammar only, and then discriminate inputs based on how many constraints they satisfy, essentially ranking inputs based on constraint satisfaction and evolving inputs to eventually obtain a set of valid inputs.
As constraints can be arbitrary Python code, our insight in this paper is to train a classifier on the private data, and use \textit{that classifier} as a constraint.
Note that Fandango can be supplied with an \textit{initial population} of inputs; we will leverage this later to supply Fandango with the best synthetic data as a starting point.

\subsection{Generative Adversarial Networks}

A generative adversarial network (GAN) is a deep learning model which incorporates two competing (adversarial) neural networks: a \textit{generator} to generate fake data and a \textit{discriminator} to spot fake data.
Generators typically start by producing noise, and that noise is gradually refined into real-looking data through feedback from the discriminator, which itself is trained on the actual data.
Since the generator is a neural network, it is generally not possible to know exactly \textit{what} properties of the dataset are learned; our idea in this paper is to replace the opaque generator with a simple, transparent, easy-to-understand data specification.

\subsection{Generating Synthetic Data}

Generating synthetic data has been extensively studied using different variations of GANs~\cite{challagundla2025synthetic}. Initially, GANs have been mostly leveraged for image generation~\cite{goodfellow2014generative}. Due to their great capabilities of generating data that is very close to the original data, they have been investigated in the context of generating synthetic tabular datasets. Xu, et al. introduced CTGAN~\cite{xu2019modeling} a conditional GAN introducing conditions in the generation process to address the mode collapse issue leading to inefficient datasets. Later, CTAB-GAN~\cite{zhao2021ctab} has been introduced to improve the results for tabular datasets containing a large amount of imbalanced categorical features. CTAB-GAN adds information loss into the conditional GAN architecture to address this challenge. While many techniques have been introduced to improve the resemblance and utility of synthetic datasets by leveraging GANs or by using statistical models~\cite{zhang2017privbayes} they tend to generate data that is to close to the original data and thus does not provide sufficient privacy guarantees. To improve privacy traditional anonymization techniques such as masking~\cite{fuller1993masking}, generalization~\cite{samarati1998generalizing} and suppression~\cite{samarati1998protecting} can be used to improve privacy models such as k-anonymity~\cite{sweeney2002k} or differential privacy~\cite{dwork2008differential}. However, current state-of-the-art techniques face a trade-off between utility and privacy which we aim to overcome with our approach.

\section{Approach}
\label{sec:approach}

\begin{figure}
    \centering
    \includegraphics[width=\linewidth]{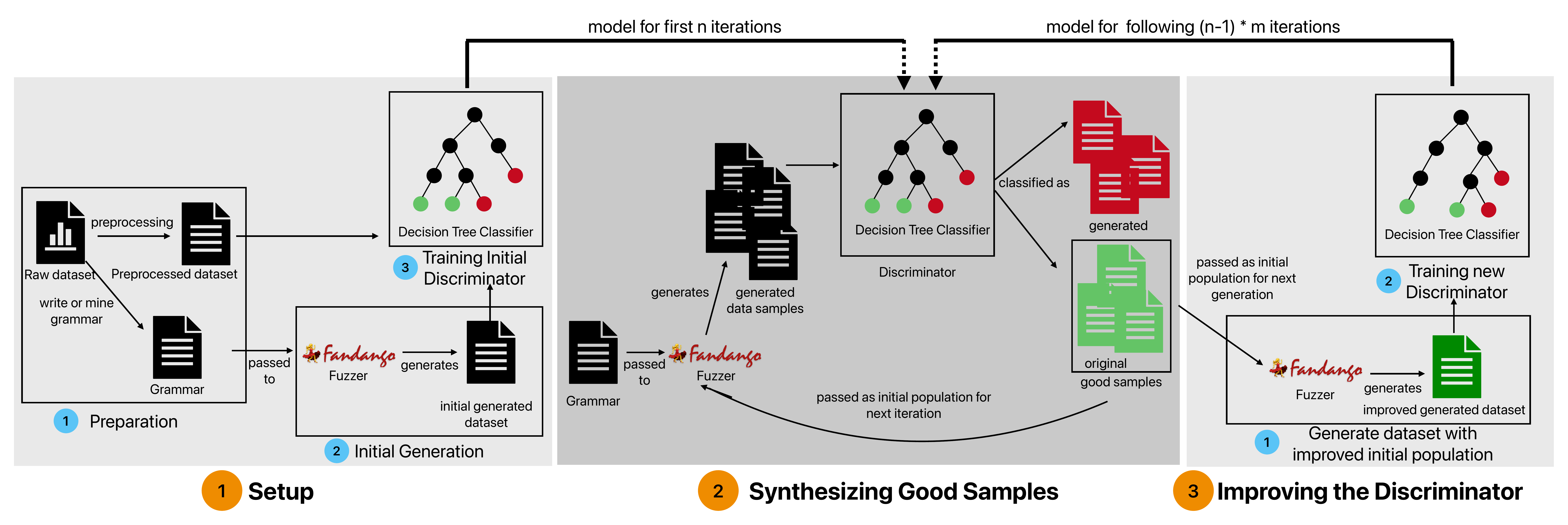}
    \caption{Overview of the approach.}
    \label{fig:approach}
\end{figure}

To generate privacy-preserving synthetic data, we introduce a novel approach combining fuzzing and machine learning in a GAN-inspired setup.
The basic idea is as follows:
\begin{itemize}
    \item First, \textit{define a Fandango specification} for the data format and generate an initial synthetic dataset.
    \item Second, \textit{train a classifier} on the original (private) and synthetic data.
    \item Finally, use that classifier as a \textit{constraint} in Fandango, satisfied if generated data is classified as original.
\end{itemize}
Figure~\ref{fig:approach} provides an overview of the setup, the generation of ``good samples'' and the retraining of the discriminator, and the remainder of this section describes each phase of the approach in turn.

\subsection{Setup (Phase 1)}

\subsubsection{Data Format Specification and Initial Generation.}
As a first step, our approach requires a Fandango specification for the data format. 
Such a specification is a mix of grammar and constraints; see Figure~\ref{fig:background:example-grammar} for an example.
This is a manual step, but specifications for tabular data are quite simple.
While Fandango does give the option to define constraints on the data, we found this is not so necessary as the discriminator handles data constraints implicitly when classifying data over a few iterations.
Fandango is then invoked to generate an initial synthetic dataset comprising as many samples as the original dataset.

Note, we perform some basic pre-processing on the original data, such as removing duplicates, removing rows with missing values, and converting categorical values to integers to convert each column in the dataset into a type that can be passed to machine learning models.
Crucially, though, \textit{Fandango never sees the private data}.

\subsubsection{Train Classifier on Private Data.}
Our approach requires to train at least one discriminator on the initial synthetic data and original private data.
In this paper, we chose to leverage three classifiers during this step since different classifiers might be able to capture different relationships of the samples and thus predict different samples as good samples. A decision tree classifier, random forest and k nearest neighbors were trained as discriminators, where 80\% of the data was used for training and 20\% for testing. In principle, any kind of classifier can be used as discriminator and easily added to the model registry in our tool.
In this initial step, all discriminators can very accurately distinguish between original and generated data (96-98\% accuracy in our experiments). To establish a set of good samples to be used as initial population, we adopted a dual approach to maximize diversity and quality. To keep only good samples that have a good chance to be representative in this initial step, we only keep a generated samples if at least two of the discriminators consider it a good sample, if only a single discriminator considers it as good we discard the sample. To maximize diversity, we repeat this process ten times. The discriminator with the best accuracy on the test set will be used in the next phase.

\subsection{Synthesizing Good Samples (Phase 2)}
In the second phase of our approach, the goal is to have Fandango generate data samples that fool the discriminator and are ``close'' to the original data.
Iteratively, Fandango will generate samples and query the discriminator to see if the generated samples are classified as original.
Every sample that fools the discriminator in this way is considered a ``good sample'', and these good samples are added to Fandango's initial population in subsequent iterations.
Fandango works by evolving inputs by mutation and crossover to try to satisfy constraints, so an initial population serves to focus Fandango on promising inputs---otherwise, it will generate samples according to the grammar.
Iterations continue until no new good samples are found, and when the accuracy of the discriminator decreases to random guessing (around 50\%) or when as many good samples as the original dataset size are found --- we then train a new discriminator.

\subsection{Improving the Discriminator (Phase 3)}
We train a new discriminator on the original data \textit{and the good data generated by Fandango} in the previous loop. 
Retrained discriminators have slightly lower accuracy, but are still accurate enough to detect differences between generated and original samples, and so the new discriminator is passed to Phase 2 to collect a new set of good samples.
This process of retraining a new discriminator is repeated until the synthetic data provides a satisfying utility, referring to how well a specific task, e.g., predicting insurance charges, can be performed with the synthetic dataset compared to the original dataset.

\subsection{Summary of Approach}

A Fandango specification describes the input format, and a discriminator serves as a constraint: generated samples that fool the discriminator are considered to have satisfied the constraint.
Initially, the discriminator is trained to differentiate purely random tabular data and the original data.
Then, the specification and discriminator are used to generate synthetic data that passes as original (called \textit{good samples}), and once we have enough data, we train a more discerning discriminator to differentiate the good data from the original, and then generate better synthetic data.
This process repeats until the synthetic data has high utility for some task.

\section{Evaluation} 
\label{sec:eval}

The goal of our experiments is to investigate the potential of generating privacy-preserving synthetic tabular datasets with a classical input generation approach. 
We aim at generating synthetic datasets with similar \textit{utility} to the original dataset, with high \textit{statistical resemblance} while \textit{providing privacy}. 

\subsection{Evaluation Setup}
\label{sec:eval:plan}

In our evaluation, we pose and answer the following research questions:

\begin{itemize}
    \item \textbf{RQ1}: How \textit{feasible} is it to generate good samples with Fandango?
    \item \textbf{RQ2}: How \textit{effective} are datasets generated by our tool?      
    \item \textbf{RQ3}: How well does the generated data \textit{match} the original data, and does it preserve privacy?
\end{itemize}

\begingroup
\setlength{\tabcolsep}{4pt}
\begin{table}[b]
    \centering
    \caption{Summary of datasets}
    \begin{tabular}{lllrr}
    Dataset & Domain & Description & Samples & Features \\
    \hline
    \texttt{heart}~\cite{heart_disease_45} & Health & Predict heart disease (classification) & 303 & 13 \\ 
    \texttt{mpg}~\cite{auto_mpg_9} & Industrial & Predict miles per gallon (regression) & 398 & 7 \\
    \texttt{grades}~\cite{student_performance_320} & Education & Predict grade (regression) & 649 & 30 \\
    \texttt{insurance}~\cite{insurance_kaggle} & Finance & Predict charges (regression) & 1338 & 7 \\
    \texttt{bank}~\cite{bank_marketing_222} & Finance & Predict term deposit (classification) & 4521 & 17 \\
    \texttt{law school}~\cite{law_school_kaggle} & Education & Predict bar exam pass (classification) & 22407 & 39 \\ 
    \texttt{adult}~\cite{adult_2} & Finance & Predict income (classification) & 32562 & 15 \\
    \texttt{credit}~\cite{credit_kaggle} & Finance & Predict loan status (classification) & 32581 & 12 \\
    \end{tabular}

    \label{tab:datasets}
\end{table}
\endgroup

\paragraph*{Datasets} Table~\ref{tab:datasets} summarizes the datasets used in our evaluation which we chose for:
\begin{itemize}
    \item \texttt{adult} and \texttt{bank} are from the UCI dataset database, and represent relatively simple and common datasets with a lot of samples for \texttt{adult};
    \item \texttt{credit} and \texttt{insurance} are from Kaggle, representing again relatively simple and common datasets;
    \item \texttt{law school} is also from Kaggle, and is a dataset often used to investigate machine learning fairness~\cite{chen2024fairness}, and has many features;
    \item \texttt{heart} is from UCI, selected as it has relatively few samples;
    \item \texttt{grades}~\cite{student_performance_320} is a dataset from UCI, selected due to the regression task and relatively few samples (we use the Portuguese grade dataset, not the math one);
    \item \texttt{mpg}~\cite{auto_mpg_9} is our final dataset, also from UCI, selected due to the regression task as well as having even fewer samples and features.
\end{itemize}

\paragraph*{Definitions \& Evaluation Metrics} To evaluate effectiveness, we need to measure the resemblance, utility, and privacy. 
To avoid ambiguity, consider the following definitions and evaluation metrics:
\begin{itemize}
    \item \texttt{Resemblance}: To assess the statistical similarity, we use three different metrics. We use Jensen-Shannon divergence~\cite{lin1991divergence} and Wasserstein distance~\cite{ruschendorf1985wasserstein} to measure the difference between the probability distributions of categorical and continuous variables in the original and generated datasets, respectively. 
    \item \texttt{Utility}: The utility is assessed by comparing the performance of a model trained and tested on the original dataset and a model trained on the generated data but tested on the original dataset. If they perform similarly, the synthetic dataset has high utility and could replace the original dataset for privacy preserving reasons while providing the same (or similar) efficiency for the task. The performance of the model will be measured by the accuracy for classification tasks and with the R2 score for regression tasks. 
    \item \texttt{Privacy}: The privacy is first assessed by verifying that no generated samples were part of the original dataset. Second, we perform a feature ablation study and check until when a sample remains \textit{unlinkable} to the original dataset.
\end{itemize}

\subsection{Experimental Setup}
In this study, we wrote Fandango specifications for each dataset to ensure their correctness. 
(Note that this could be automated in the future by using grammar or specification mining techniques~\cite{10.1145/3776743,liu2025synthesizing}.) 
The selected datasets cover classification as well as regression tasks. 
To determine the utility of the original and synthetic datasets, several models were trained. 
All implementation has been done in Python using the scikit-learn\footnote{\url{https://scikit-learn.org/stable/}} library. 
The dataset has been split into 80\% for training and 20\% for testing in all the experiments.

Our approach contains a few variables in the configuration that can be adjusted depending on the dataset size and the desired experiments. 
Notably, the number of desired solutions and the number of maximum iterations performed by Fandango to collect the good samples have, in all experiments, been set to the dataset size. 
However, the maximum number of iterations was never required in our experiments since more than one good sample was found in each iteration.
The next variable in our pipeline is how often the discriminator gets retrained on the latest generated samples. 
This part remains variable in the subsequent experiments and will be explicitly stated for each one.

To answer \textbf{RQ1}, we track how many Fandango-generated samples fool the discriminator, and investigate how many iterations are required before retraining the discriminator. Additionally we report the average generation time required to get a synthetic dataset composed of as many good samples as the original dataset size.
For \textbf{RQ2}, we need to evaluate the utility of the synthetic dataset. 
A model trained on synthetic data will have \textit{high utility} if it performs roughly as well as a model trained on the original data.
Thus, we train several models on the original and the generated datasets to perform a dedicated task. 
For the \texttt{insurance, grades}, and \texttt{mpg} dataset, we train regression models to predict the ``charges'', ``G3'', and ``mpg''. 
For the remaining datasets, we train classification models for dataset-specific prediction tasks, predicting:``income'' for the \texttt{adult} dataset, ``loan status'' for the \texttt{credit} dataset, a binary ``term deposit'' for the \texttt{bank} dataset, ``bar\_passed'' for the \texttt{la w school} dataset, and ``num'' for the \texttt{heart} dataset. 
Finally, for \textbf{RQ3} we evaluate how well our approach generates similar and privacy-preserving synthetic data; we measure the Wasserstein distance after normalizing all the features in both datasets, and the Jensen-Shannon divergence.

\emph{System Requirements} All experiments were performed on a MacBook Pro M5 Max with 128GB of RAM (but only 7GB are required to run the largest experiments). 

\subsection{Results}
\label{sec:eval:prelim}

\subsubsection{RQ1: Feasibility of generating good samples}

Table~\ref{tab:goodsamples_iter_time} illustrates the number of iterations and time required to generate the good samples. For every dataset, we can observe the same trend: the amount of good samples first increases rapidly and later increases more slowly until the desired amount of good samples is reached.
The evolutionary algorithm is not deterministic; in some runs the mutations might directly be performed on the most interesting parts of the samples, while in other runs it might take longer to guide the mutations towards better samples. 
Consequently, we performed the experiment ten times and reported the average. 

Overall, the number of iterations required to get a synthetic dataset of the same size as the original dataset, containing only good samples, varies between 6 iterations for the smallest dataset and up to 87 iterations for the large one. 
We see that the time required to generate the desired amount of good samples scales proportionally to the dataset size. 
The number of iterations required represents in the worst case 2\% of the dataset size.
The time and iteration number also depend on the accuracy of the discriminator and the quality of the grammar. 
The more general the grammar, the longer it will take to evolve the samples towards ``good samples'' that capture the correlation between the variables. 
Nevertheless, it is feasible to generate sufficiently many samples that fool a discriminator using classical input generation techniques. 

\begingroup
\setlength{\tabcolsep}{7pt}
\begin{table}[b]
    \centering
    \caption{Iterations and time required to collect good samples for each dataset on average}
    \begin{tabular}{lrrr}
        \textbf{Dataset} & \textbf{Good Samples (= Dataset Size)} & \textbf{Iterations }& \textbf{Time (seconds)} \\
        \hline
        \texttt{heart} & 303 & 6.3 & 4.10 \\
        \texttt{mpg} & 398 & 6.2 & 6.67\\
        \texttt{grades} & 649 & 5.6 & 12.64\\
        \texttt{insurance} & 1338 & 7.4 & 10.53 \\
        \texttt{bank} & 4521 & 19.8& 162.89\\
        \texttt{law school} & 22407 & 51.5 & 8513.42 \\
        \texttt{adult} & 32562 & 86.9 & 7651.05\\
        \texttt{credit} & 32581 & 53 & 7888.34\\
    \end{tabular}
    \label{tab:goodsamples_iter_time}
\end{table}
\endgroup

\subsubsection{\textbf{RQ2}: Utility of Synthetic Data}

In this research question, we investigate the utility of the following combinations:
\begin{itemize}
    \item For the task: \textbf{Train Original - Test Original}, the model is trained and tested on the original dataset to establish a baseline of how ``good'' the original data is.
    \item For the task: \textbf{Train Generated - Test Generated}, the model is trained and tested on the generated dataset to evaluate consistency in the generation pattern.
    \item For the task: \textbf{Train Original - Test Generated}, the model is trained on original data and tested on the generated dataset to assess how well AI models can be tested and validated with synthetic data.
    \item For the task: \textbf{Train Generated - Test Original}, the model is trained on generated data and tested on the original dataset to assess if models can be trained on synthetic datasets to preserve privacy while keeping similar utility as trained with original data.
\end{itemize}

Table~\ref{tab:utility_regression} illustrates the utility of the synthetic datasets used for \texttt{regression} tasks.
The baseline models trained on the original dataset reach R2 scores between 0.83 and 0.91.
A preliminary study testing several model architectures across all datasets revealed that the \texttt{RandomForestRegressor} and the \texttt{GradientBoostingRegressor} performed best. Thus, for all regression tasks both models were trained and tested. 

In Table~\ref{tab:utility_regression}, the score of the best performing model is reported. Across all datasets, the R2 score for the task \emph{Train Generated - Test Generated} is the highest, suggesting that all the samples in the synthetic dataset have been evolved using the same patterns and that the relationship between the features and label has been well replicated. \emph{Train Generated - Test Original} is particularly relevant to assess whether a model can be trained on a synthetic dataset to avoid data leakage while providing similar utility. 
For the \texttt{insurance} dataset, the synthetic dataset can be used for that purpose. For \texttt{grades} and \texttt{mpg} further investigations would be required for good utility. The \texttt{grades} dataset is particularly challenging for our approach since it comes with relatively few samples but there are many features (see Table~\ref{tab:datasets}). 
It is likely that our approach would require more retraining loops to better capture the complex relationships between the variables. \emph{Train Original - Test Generated} demonstrates the potential of our approach to test and validate models trained on sensitive data, where the original dataset cannot be shared for testing and validation purposes. This task was challenging on the \texttt{mpg} datasets where the label can take a wide range of values. Nevertheless, for \texttt{insurance} and \texttt{grades} acceptable utilities were reached.

\begingroup
\setlength{\tabcolsep}{10pt}
\begin{table}[b]
    \centering
    \caption{Utility (R2 score) of the Synthetic Datasets for Regression Tasks}
    \begin{tabular}{lrrrr}
        \textbf{Task} &  \textbf{insurance} & \textbf{grades} & \textbf{mpg} \\
        \hline
        Train Original - Test Original & 0.8779  & 0.8394 & 0.9138\\
        Train Generated - Test Generated & 0.9388  & 0.9680 & 0.9031\\
        Train Generated - Test Original & 0.7525 & 0.2670 & 0.5538 \\
        Train Original - Test Generated & 0.6878  & 0.7722 &  0.2401\\
    \end{tabular}
    \label{tab:utility_regression}
\end{table}
\endgroup

In Table~\ref{tab:utility_classification}, the baseline accuracy of the original data is reported for  each dataset used for \texttt{classification} tasks as well as the accuracy of the synthetic datasets. Similar to regression tasks, the baseline reaches between 85\% and 100\% accuracy.
As we observed for the regression task, training and testing on the same datasets works quite well, and utility drops when training on one dataset, and testing on the other. 
Comparing Tables~\ref{tab:utility_regression} and~\ref{tab:utility_classification}, cross-dataset results are better for classification than regression, on average.
We hypothesize that this is due to the relatively more restricted set of possible results for classification, and the odds of Fandango generating unrealistic classification target values is much lower than the odds of generating unrealistic regression target values.

\begingroup
\setlength{\tabcolsep}{7pt}
\begin{table}[t]
    \centering
    \caption{Utility (accuracy) of the Synthetic Datasets for Classification Tasks}
    \begin{tabular}{lrrrrr}
        \textbf{Task} &  \textbf{adult} & \textbf{bank} & \textbf{credit} & \textbf{law school} & \textbf{heart} \\ 
        \hline
        Train Original - Test Original & 0.8521 & 0.8972 & 0.9305 & 1.0 & 0.8524 \\ 
        Train Generated - Test Generated & 0.8895 & 0.9381 & 0.8587 & 0.9615 & 0.9180 \\ 
        Train Generated - Test Original & 0.7758 & 0.7779 & 0.6921 & 0.9003 & 0.6229 \\ 
        Train Original - Test Generated & 0.6058 & 0.6552 & 0.5888 & 0.6108 & 0.6393\\
    \end{tabular}
    \label{tab:utility_classification}
\end{table}
\endgroup

On the whole, for half of the datasets, the score reached while training and testing on synthetic datasets were even higher than the baseline scores. 
This is likely because the discriminator provided valuable feedback for Fandango to learn inter-feature dependencies, leaving little room for outliers to be present in the generated dataset. 
When evaluating the classifiers trained on original and tested on generated data (and vice versa), accuracy is slightly lower but remains satisfying, showing the same utility trend as synthetic datasets used for regression tasks.

Overall, our experiment has shown that it is feasible to generate synthetic datasets with similar utility as the original dataset. 
\subsubsection{\textbf{RQ3}: Similarity and Privacy of Synthetic Data}
To assess how well the synthetic datasets match the original dataset in terms of statistical similarity while preserving privacy, two experiments were conducted.

\paragraph{Statistical similarity} A first experiment investigated how close the synthetic datasets with the best utility are to the original data. To assess the statistical similarity the Wasserstein distance was computed for the continuous variables while the Jensen Shannon divergence was computed for categorical variables. The values reported in Table~\ref{tab:similarity_metrics} represent the median of all the respective variables. Values close to zero indicate that the original and synthetic dataset are very similar while values close to one highlight that both datasets are statistically very different. 

For half of the synthetic datasets, the Wasserstein distance was smaller or equal to 0.11 which indicates that both the original and generated datasets are very similar. 
Three more datasets had distances between 0.14 and 0.18, which are still reasonable, and such distances indicate the statistical similarity while preserving some distance to the original dataset which increases the privacy. The largest distance was observed for the \texttt{law school} dataset, which is the dataset containing the most features. Thus, it is likely that larger distances remained since not all the features are very important for the given task. Thus, the Fandango fuzzer might happily mutate those parts of the samples, not impacting the utility but increasing the distance from the original samples.
Recent literature~\cite{challagundla2025synthetic} provided some insights into the Jensen Shannon divergence and Wasserstein distance of the \texttt{Adult} and \texttt{Credit} with the current state-of-the-art techniques. 
Techniques like CTGAN~\cite{xu2019modeling} achieve very low distances as the generated data is very close to the original data, leading to good utility but making them vulnerable to several attacks. 
Thus, a small Wasserstein distance or low Jensen-Shannon divergence does not imply strong privacy guarantees. 
With our approach, we achieve in a few cases a slightly higher but still acceptable Wasserstein distance. Several GAN-based techniques evaluated in~\cite{challagundla2025synthetic} reached Wasserstein distances between 0.03 and 0.25 with an accuracy on the bank and credit dataset varying between 40\% and slightly above 80\%. 

When comparing the Wasserstein distance and the Jensen Shannon divergence, overall the scores for Jensen Shannon remain slightly higher. However, this is a common trend and those to different metrics cannot be compare one on one. Nevertheless, the Jensen Shannon divergence remains very low for the \texttt{insurance} and \texttt{grades} datasets which also achieve small Wasserstein distances. Thus, those synthetic datasets have overall a good statistical similarity. Also for \texttt{bank} and \texttt{credit}, a statistical similarity can be observed. However, for \texttt{mpg} and \texttt{law school} no statistical similarity can be obeserved for the categorical variables. In the case of the \texttt{law school} dataset, as for the Wasserstein distance, the large diversity of features makes it far more challenging to keep a close similarity for all variables. For \texttt{mpg}, only two features where categorical features which might be less representative for the learning, and thus less focused on during the guided generation.

\begingroup
\setlength{\tabcolsep}{10pt}
\begin{table}[t]
    \centering
    \caption{Similarity Metrics for every Dataset}
    \begin{tabular}{lrr}
         \textbf{Dataset} &  \textbf{Wasserstein distance} & \textbf{Jensen Shannon}\\
         \hline
         \texttt{heart} & 0.17 & 0.38\\
         \texttt{mpg} & 0.14 & 0.77\\
         \texttt{grades} & 0.18 & 0.17\\
        \texttt{insurance} & 0.11 & 0.02\\
        \texttt{bank} & 0.10 & 0.25\\
        \texttt{law school} & 0.41 & 0.44 \\
        \texttt{adult} & 0.11 & 0.22\\
        \texttt{credit} & 0.10 & 0.29\\
    \end{tabular}
    \label{tab:similarity_metrics}
\end{table}
\endgroup

\paragraph{Unlinkability \& Feature Ablation Study} 
A second experiment was conducted to evaluate how well the synthetic data preserves the privacy of the original dataset. 
While the generator has never seen the original private dataset, potential private information flow through the discriminator's feedback is still possible. 
Thus, a first check verified that no generated sample was part of the original dataset, and this was true for all datasets.

Next, a feature ablation study was performed. 
It is important to assess if generated samples are equal to original samples \textit{modulo a few features}. 
Thus, we remove one feature after another to verify if partial samples match the original samples. In addition, we investigated which features distinguish the most between the original and generated samples. 
A hypothesis is that some features are less important when training a model for a given task, thus changing the value of such a variable might not influence wether a sample is considered a good sample or not. 
Due to the fact that the Fandango fuzzer uses an evolutionary algorithm and mutates good samples, it could be that several similar good samples are generated. 
In Table~\ref{tab:feature_ablation} we present for each dataset (1) the most frequent variables that, if removed, yield partial samples matching some original samples, (2) for how many samples this is the case (in percentage to allow a fair comparison between datasets of varying sizes), and (3) the impact of removing this feature while training the model for the given task.

\begingroup
\setlength{\tabcolsep}{4pt}
\begin{table}[!h]
    \centering
    \caption{Feature ablation study. 
    Score is the accuracy or R2 score of the original dataset.
    The \texttt{law school} dataset had only two columns where, if removed, we saw partial matches with the private data, which we report here.
    For \texttt{credit}, we did not ablate the \texttt{loan\_status} feature since it is the target label for prediction.}
    \begin{tabular}{llrllr}
          &  &  & \multicolumn{3}{l}{\textbf{Top 3 features relevant for unlinkability,}} \\
         \textbf{Dataset} & \textbf{Score} & \textbf{$\#$ Features} & \multicolumn{3}{l}{\textbf{and impact on model training}} \\
         \hline
         \texttt{insurance} & 0.8779 & 7 & \texttt{region} & 0.7\% & 0.00 \\
         & & & \texttt{smoker} & 0.5\% & -0.83 \\
         & & & \texttt{children} & 0.5\% & -0.01 \\
         \texttt{bank} & 0.8972 & 17 & \texttt{pdays} & 1.1\% & 0.00 \\
         & & & \texttt{duration} & 0.5\% & -0.07 \\
         & & & \texttt{campaign} & 0.2\% & 0.01 \\
         \texttt{credit} & 0.9305 & 12 & \texttt{loan\_int\_rate} & 2.8\% & 0.00 \\
         & & & \texttt{loan\_status} & 2.3\% & -- \\
         & & & \texttt{loan\_percent\_income} & 0.2\% & 0.00 \\
         \texttt{adult} & 0.8521 & 15 & \texttt{capital\_loss} & 1.1\% & 0.00 \\
         & & & \texttt{capital\_gain} & 0.7\% & -0.04 \\
         & & & \texttt{sex} & 0.2\% & 0.00 \\
         \texttt{law school} & 1.0000 & 39 & \texttt{dnn\_bar\_pass\_prediction} & 0.07\% & 0.00 \\
         & & & \texttt{gpa} & 0.004\% & 0.00 \\
         \texttt{heart} & 0.8524 & 13 & \texttt{oldpeak} & 7.2\% & 0.06 \\
         & & & \texttt{slope} & 6.2\% & -0.1 \\
         & & & \texttt{ca} & 2.3\% & 0.03 \\
         \texttt{grades} & 0.8394 & 30 & \texttt{G1} & 2.0\% & 0.02 \\
         & & & \texttt{G2} & 1.7\% & -0.16 \\
         & & & \texttt{absences} & 0.9\% & -0.01 \\
         \texttt{mpg} & 0.9138 & 7 & \texttt{origin} & 5.7\% & 0.00 \\
         & & & \texttt{year} & 3.8\% & -0.11 \\
         & & & \texttt{acceleration} & 1.0\% & -0.02 \\
    \end{tabular}
    \label{tab:feature_ablation}
\end{table}
\endgroup

As can be seen in Table~\ref{tab:feature_ablation}, in most cases the top 3 features that, if removed lead to partial matches with original samples, are features that have very little impact on the models performance when training with and without that feature. 
Only a single case was observed for the \texttt{insurance} dataset, where a few partial matches were found when removing the \texttt{smoker} feature, even though this feature contributes most to the model training. 
However, this was only the case for 0.5\% of the samples. 
For the \texttt{grades} and the \texttt{mpg} dataset, one feature was detected for a few samples, which also has some impact on the models training. 

An interesting finding is that for the \texttt{credit} dataset, some partial matches were obtained by removing the label (i.e., the prediction target), meaning that for 2.3\% of the synthetic dataset, the generated label was actually in contradiction with the original label. 
There might be a subset of samples that are very close to the classification boundary, leading to fandango generating samples with a contradictory label which are still considered as correct by the discriminator. 

Overall, for all the other datasets, the commonly identified features that when removed can lead to partial matches have no impact on the models training. 
We believe that this is because the mutations that the fuzzer makes which do preserve the ``goodness'' of a sample tend to be on the features that are less important for utility.
Recall that Fandango is an evolutionary language-based fuzzer, and the two major mechanisms for mutating inputs are individual mutation and cross-over; the discriminator must have internalized the relationship between features and label (as evidenced by the overall high score of training and testing on generated data in RQ2), which leads it to reject samples that differ too much in important features.

Regarding \emph{privacy}, no direct matches were generated with our approach and only a small subset of partial matches were detected. 
For the smallest dataset (\texttt{heart}), 7.2\% of generated samples were partial matches, the highest proportion overall. 
This is likely due to \texttt{heart} being a very small dataset, and only a small amount of generations were performed by Fandango. 
Additionally, \texttt{mpg}, also very small, has only 7 features which increases the likelihood of partial matches. 
Further, only the top-3 features were reported since the amount of partial matches become negligible for the remaining features with no partial matches for most of them. 
This trend was also observed for the \texttt{law school} dataset where only for two columns, if removed, we saw partial matches with the original dataset.

\subsubsection{Summary of Findings}
In our experiments using SecretFan to generate synthetic datasets, we have found in RQ1 that generating synthetic datasets can be done in a reasonable time on commodity hardware. In RQ2 we evaluated the utility of the synthetic datasets. Overall, the models trained and tested on the generated datasets, all had good accuracy and r2 scores. While there were some variations for the models trained on generated data and tested on original and vice versa, for most datasets we still achieve acceptable utility compared to the baseline. Additionally, we have good similarity scores in terms of Wasserstein distance and Jensen Shannon Divergence, and relatively few samples are partial matches to the originals, with no direct matches.
Overall, we believe that our empirical results are encouraging when it comes to generating privacy-preserving synthetic datasets. This work also opens many new research directions using traditional input generation techniques towards generating privacy preserving synthetic datasets.

\section{Discussion}

While we believe our work shows promising results, there are many direct follow-up questions that are worth investigating.

\paragraph{Investigating impact of specification detail}

As an example, there are many ways to express an \textit{age} in a Fandango specification:
\begin{enumerate}
    \item \texttt{<age> ::= <digit>+ }
    \item \texttt{<age> ::= <non\_zero> <digit>}
    \item \texttt{<age> ::= <digit>+}, with a constraint \texttt{where 18 <= int(<age>) <= 78}
\end{enumerate}

A discriminator should be able to internalize what a reasonable age is for a given dataset by being trained on it, but being more specific can help an evolutionary input generator like Fandango to be more efficient and generate good samples more constructively.
We were judicious when writing the specifications, as a truly accurate specification would require some domain knowledge of the dataset, though we envision that organizations wanting synthetic data for their private datasets would have deep domain knowledge of their own data.

\paragraph{Adding more targeted constraints}

The Wasserstein distance and Jensen Shannon divergence metrics are a way to assess the similarity of two populations.
While we computed these post facto, as part of the evaluation, we hypothesize that they could be directly included \textit{in a Fandango spec as constraints}.
Then, the evolutionary algorithm would try to maximize distance scores while also maximizing utility for the whole population, or minimize overall distance while requiring that no individual samples are too close to private ones (in terms of partial matches).
This would require an extension to Fandango for it to support population-level constraints.

\subsection{Limitations}
Generating synthetic data with Fandango requires a Fandango specification. Such a specification can be manually written or inferred from samples or natural language descriptions. However, this assumes that either a set of samples or a clear description of the data is available.
The use case we envision for this work is that of an organization wanting to make a synthetic version of their private data available to the public, and we believe such an organization will be composed of domain experts about the data, so they should have sufficient knowledge to prepare good specifications.
For our evaluation, we were judicious in our choice of which constraints to add and how to express the grammar, basing ourselves on the publicly available descriptions of the data sets.

\subsection{Threats to Validity}

In our evaluation, we used eight commonly used datasets, where five datasets are used for a classification tasks and three for regression tasks.
To evaluate the resemblance of the synthetic and original datasets, we have computed the Wasserstein distance and the Jensen Shannon Divergence for the synthetic datasets with the best utility. 
However, we have not further tracked the evolution of the statistical similarity during the generation and retraining process. 
In the future, the information could be leveraged during the generation process.

To assess the privacy, a membership and feature ablation study was performed, but no attacks were performed on the models trained on synthetic data.
The utility of the generated datasets varies based on the ``good samples'' collection. 
In our implementation, we considered a sample as good and used it in the initial population if at least two discriminators classified it as original. 
However, focusing on quality might slightly reduce the diversity of samples. 
Beyond this, we could use discriminators differently, e.g., classifying a sample as ``good'' only if the discriminator predicts it as such with high confidence. 
We consider this an interesting avenue for near-term future work, as the aim of this work is to provide a high-level alternative solution to generate privacy preserving synthetic dataset with similar utility than the original one. 
Additionally, with our generator, it can be observed how the samples are evolved, adding to the explainability of how the dataset was generated and which features the generator focused on, which would also be interesting to investigate further.

\section{Conclusion and Future Work}
\label{sec:conclusion-and-future-work}

Training AI systems on real-world data is effective, but it bears the risk of the model memorizing and replicating private or confidential information.
How can we produce synthetic data that replicates essential features of original data without memorizing the original data?
We have presented a novel approach based on language-based testing:
\begin{enumerate}
\item An \textit{input generator} produces synthetic data from an input specification, possibly replicating constraints of the original dataset
\item A \textit{discriminator} trained from the original data guides the evolution of the synthetic data towards the features of the original dataset
\end{enumerate}
By construction, this ensures there is no \emph{data flow} from the original data to the synthesized data.
The amount of \textit{information flow} from the original data can be controlled through the \textit{strictness} of the discriminator: If the discriminator accepts \textit{only} the original data, then obviously any synthesized data will be identical.
A less overfitting discriminator, though, will result in synthesized data that shows a \textit{balance} between privacy and utility; our experimental results indicate that our approach can actually achieve such a balance.

In the long run, we believe this work forms the foundation for many exciting future research directions:

\begin{description}
\item[Tunable privacy.] We plan to integrate discriminator guidance right into Fandango constraints, such that users can themselves specify how close the synthetic data should be to the original data.
\item[On-demand generation of synthetic data.] While Fandango constraints are mostly concerned with semantic correctness for having inputs get accepted as valid, we plan to devise dataset-level constraints that focus on statistical features, further enhancing the quality of the generated data.
\item[Explainability.] We envision that our generation technique can provide benefits beyond good utility and privacy: in every generation step, one could investigate the relative importance of dataset features, as well as which samples are partial matches.
We imagine that this could be leveraged to make synthetic data generation more \textit{explainable}, as at every step the influence of dataset features on the quality and privacy of the generated data is clear.
This is in stark contrast to GAN techniques, in which the generator is a black box, and a benefit of a transparent classical technique.
\end{description}

All code and experimental data of our approach are available as open source at \texttt{\url{https://github.com/LaPlei96/SecretFan}} and the Fandango fuzzer can be found at \texttt{\url{https://Fandango-fuzzer.github.io/}}.

\bibliographystyle{splncs04}
\bibliography{references}

\end{document}